\documentclass{ceurart}

\usepackage{listings}  
\usepackage{algorithm}      
\usepackage{algpseudocode}
\usepackage{graphicx} 
\usepackage{subcaption}
\usepackage{todonotes}
\usepackage{pifont}
\usepackage{soul}
\usepackage{makecell}
\usepackage{fontspec}
\usepackage{kotex}
\newfontfamily\telugufont{NotoSansTelugu}[
Path=fonts/NotoSansTelugu/, Extension = .ttf, UprightFont = *-Regular, BoldFont = *-Bold]
\newfontfamily\bengalifont{NotoSansBengali}[
Path=fonts/NotoSansBengali/, Extension = .ttf, UprightFont = *-Regular, BoldFont = *-Bold]
\begin{document}

\copyrightyear{2025}
\copyrightclause{Copyright for this paper by its authors.
  Use permitted under Creative Commons License Attribution 4.0
  International (CC BY 4.0).}

\conference{CLEF 2025 Working Notes, 9 -- 12 September 2025, Madrid, Spain}
\title{AKCIT-FN at CheckThat! 2025: Switching Fine-Tuned SLMs and LLM Prompting for Multilingual Claim Normalization}
\title[mode=sub]{Notebook for the CheckThat! Lab at CLEF 2025}

\author[1,2]{Fabrycio Leite Nakano Almada}[%
orcid=0000-0003-3876-9635,
email=fabrycio@egresso.ufg.br,
]
\fnmark[1]

\author[1,2]{Kauan Divino Pouso Mariano}[%
orcid=0009-0008-9082-0876,
email=kauan@discente.ufg.br,
]
\fnmark[1]

\author[1,2]{Maykon Adriell Dutra}[%
orcid=0009-0000-0813-3084,
email=maykonadriell@discente.ufg.br,
]
\fnmark[1]

\author[1,2]{Victor Emanuel da Silva Monteiro}[%
orcid=0009-0008-9059-6843,
email=victor_emanuel@discente.ufg.br, 
]
\fnmark[1]

\author[1,2]{Juliana Resplande Sant’Anna Gomes}[%
orcid=0000-0001-6900-1931,
email=juliana.resplande@discente.ufg.br,
]
\cormark[1]

\author[1,2]{Arlindo Rodrigues Galvão Filho}[%
orcid=0000-0003-2151-8039,
email=arlindogalvao@ufg.br,
]

\author[1,2]{Anderson da Silva Soares}[%
orcid=0000-0002-2967-6077,
email=andersonsoares@ufg.br,
]

\address[1]{Institute of Informatics, Federal University of Goiás, Brazil}
\address[2]{Advanced Knowledge Center in Immersive Technology (AKCIT), Federal University of Goiás, Brazil}

\cortext[1]{Corresponding author.} 
\fntext[1]{These authors contributed equally.}

\begin{abstract}
Claim normalization, the transformation of informal social media posts into concise, self-contained statements, is a crucial step in automated fact-checking pipelines. This paper details our submission to the CLEF-2025 CheckThat! Task~2, which challenges systems to perform claim normalization across twenty languages, divided into thirteen supervised (high-resource) and seven zero-shot (no training data) tracks.

Our approach, leveraging fine-tuned Small Language Models (SLMs) for supervised languages and Large Language Model (LLM) prompting for zero-shot scenarios, achieved podium positions (top three) in fifteen of the twenty languages. Notably, this included second-place rankings in eight languages, five of which were among the seven designated zero-shot languages, underscoring the effectiveness of our LLM-based zero-shot strategy. For Portuguese, our initial development language, our system achieved an average METEOR score of 0.5290, ranking third. All implementation artifacts, including inference, training, evaluation scripts, and prompt configurations, are publicly available at \url{https://github.com/ju-resplande/checkthat2025_normalization}.
\end{abstract}


\maketitle

\section{Introduction}

The proliferation of misinformation within social media ecosystems has intensified the demand for automated fact-checking pipelines that operate effectively across diverse languages, genres, and text characterized by significant noise. A crucial stage in such pipelines is claim normalization: the process of converting an informal, often multi-sentence social media post into a concise, self-contained statement suitable for subsequent evidence retrieval and veracity assessment. Without normalization, downstream modules must contend with extraneous elements such as redundancy, hashtags, emojis, and idiosyncratic phrasing, which collectively diminish both retrieval recall and factual accuracy \citep{sundriyal2023}.

The CLEF-2025 CheckThat! Lab confronts this bottleneck through Task 2 – Claim Normalization. Systems are required to generate normalized claims for twenty languages under two experimental conditions: (i) a monolingual setting, providing training and development splits (annotated examples) for thirteen higher-resource languages, and (ii) a zero-shot setting, releasing only test data for seven lower-resource languages, for which no specific training data is provided \citep{clef-checkthat:2025-lncs, CheckThat:ECIR2025, clef-checkthat:2025:task2}.

Our methodological development initially focused on Portuguese—the native language of our development team, facilitating a more nuanced understanding and iterative refinement—before extending validated strategies to all target languages. For languages in category (i), our experimentation involved fine-tuning open-source Encoder-Decoder Small Language Models (SLMs) and, as a comparative approach, inference via prompting with Large Language Models (LLMs). For languages in category (ii), we exclusively employed a zero-shot prompting strategy with LLMs.

Our efforts culminated in strong results, securing third place in the Portuguese subset with an average METEOR score of 0.5290. Across all languages, we achieved top-three placements in fifteen of the twenty languages. This included second-place finishes in eight languages overall (with five of these being zero-shot languages) and third-place finishes in seven languages, highlighting the robustness of our approach in both data-rich and data-scarce scenarios.

\section{Related Work}
Over the past few years, claim normalization has gained prominence as a crucial preprocessing step in automated fact-checking, moving beyond mere claim extraction. \citeauthor{Konstantinovskiy2021claim} proposed the first annotation schema for claim detection informed by experts and a benchmark for
automated claim detection that is more consistent across time, topics, and annotators than previous approaches. \citeauthor{sundriyal2023} formalized claim normalization by converting informal social-media texts into self-contained claims, demonstrating significant improvements in downstream evidence retrieval and veracity classification tasks.

Previous editions of the CLEF CheckThat! Lab (2018–2024) have included tasks such as check-worthiness detection from political debates or speeches \citep{nakov2018checkthat, tamer2019checkthat, barron2020checkthat, nakov2021checkthat}, and from political or COVID-19-related tweets \citep{barron2020checkthat, nakov2021checkthat}.

The construction of multilingual fact-checking corpora represents another active research direction. For instance, MuMiN automatically links 21 million tweets to 13 thousand fact-checked claims across 41 languages using LaBSE embeddings \citep{nielsen2022mumin}. Similarly, \citeauthor{singh2024mmtweets} created MMTweets by scraping "debunked narratives," retrieving tweets via multilingual keyword queries, and applying detailed human annotation, resulting in a multimodal dataset for cross-lingual retrieval.

\section{Task Definition}
Task 2 of the CLEF-2025 CheckThat! Lab presents a comprehensive multilingual claim normalization challenge designed to evaluate system performance across diverse linguistic contexts and resource availability scenarios. The core objective involves transforming informal social media posts into concise, self-contained, and verifiable statements while preserving all factual content and systematically removing subjective opinions, redundant expressions, and extraneous material \citep{clef-checkthat:2025-lncs, CheckThat:ECIR2025, clef-checkthat:2025:task2}.

The task employs two experimental paradigms designed to assess system robustness under varying data availability conditions:
\begin{enumerate}
   \renewcommand{\theenumi}{\alph{enumi}}
    \item \textbf{Monolingual setting:} Complete training, development, and test datasets are provided for thirteen languages: Arabic (AR), English (EN), French (FR), German (DE), Hindi (HI), Indonesian (ID), Marathi (MR), Polish (PL), Portuguese (PT), Punjabi (PA), Spanish (ES), Tamil (TA), and Thai (TH).
    \item \textbf{Zero-shot setting:} For seven languages, Only test splits are released for seven languages, requiring systems to generalize from cross-lingual knowledge or leverage multilingual pre-training: Bengali (BN), Czech (CS), Dutch (NL), Greek (EL), Korean (KO), Romanian (RO), and Telugu (TE).
\end{enumerate}

Table~\ref{tab:normalization-example} illustrates the complexity of this transformation through an English example that demonstrates the typical challenges: redundancy removal, formalization of informal language, and extraction of verifiable factual content from noisy social media text.

\begin{table}[ht]
  \caption{Example of claim normalization from an informal social media post (English dataset), demonstrating redundancy removal, formalization, and extraction of verifiable facts.}
  \label{tab:normalization-example}
  \begin{tabular}{p{8cm} p{6cm}}
    \toprule
    \textbf{Original Post} & \textbf{Normalized Claim} \\
    \midrule
    \small I guess the left is okay with this I guess the left is okay with this I guess the left is okay with this Dr. Rachel Levine @DrRachel Levine Thank you Vanity Fair for honoring me on the cover of your magazine this March. My dream of becoming @POTUS one day just took a step forward. THE SKY THE LIMIT. "Madam President Levine A LEADER IN THE MAKING 8:12 AM Feb 1, 2021 Twitter Web App ... . & \small US assistant health secretary Rachel Levine appears on the cover of Vanity Fair's March 2021 issue \\
    \bottomrule
  \end{tabular}
\end{table}

Table~\ref{tab:splits} provides a comprehensive overview of dataset statistics, revealing significant variation in resource availability across languages. This heterogeneity presents both opportunities and challenges: while resource-rich languages like English provide substantial training data (11,374 examples), smaller datasets for languages like Tamil (102 training examples) require careful consideration of overfitting risks and generalization strategies.

\begin{table}[ht]
\centering
\caption{Dataset statistics for Task 2: Claim Normalization, showing sample distribution across languages and splits.}
\label{tab:splits}

\begin{subtable}[t]{0.48\textwidth}
\centering
\caption{Monolingual setting}
\begin{tabular}{lrrr}
\toprule
\textbf{Language} & \textbf{Train} & \textbf{Dev} & \textbf{Test} \\
\midrule
English    & 11374 & 1171 & 1285 \\
Spanish    & 3458  & 439  & 439  \\
Portuguese & 1735  & 223  & 225  \\
French     & 1174  & 147  & 148  \\
Hindi      & 1081  & 50   & 100  \\
Indonesian & 540   & 137  & 100  \\
Arabic     & 470   & 118  & 100  \\
Punjabi    & 445   & 50   & 100  \\
German     & 386   & 101  & 100  \\
Thai       & 244   & 61   & 100  \\
Polish     & 163   & 41   & 100  \\
Marathi    & 137   & 50   & 100  \\
Tamil      & 102   & 50   & 100  \\
\bottomrule
\end{tabular}
\end{subtable}
\hfill
\begin{subtable}[t]{0.51\textwidth}
\centering
\caption{Zero-shot setting}
\begin{tabular}{lc}
\toprule
\textbf{Language} & \textbf{Test} \\
\midrule
Korean     & 274  \\
Dutch      & 177  \\
Greek      & 156  \\
Romanian   & 141  \\
Czech      & 123  \\
Telugu     & 116  \\
Bengali    & 81   \\
\bottomrule
\end{tabular}
\end{subtable}

\end{table}

Participant submissions were evaluated using the METEOR score \cite{banerjee-lavie-2005-meteor}, a metric commonly employed for machine translation evaluation that assesses translation quality by aligning system output with reference texts based on exact word matches, stemming, and synonymy. For this task, official scores were calculated by averaging METEOR results across all test examples for a given language/setting. Punctuation was removed during pre-processing for the evaluation script to standardize inputs and mitigate its impact on scores.


\section{Methodology}
\label{sec:methodology}
To address the claim normalization task, we employed a dual-strategy approach tailored to the different experimental settings. For the monolingual setting, where training, development, and test splits were available, we investigated both: (i) fine-tuning of open-source Encoder-Decoder Small Language Models (SLMs), and (ii) inference using Large Language Models (LLMs) with few-shot prompting. For the zero-shot setting, characterized by the absence of training data for specific languages, our efforts exclusively focused on zero-shot inference with LLMs (using prompts without in-context examples specific to the task for those languages).

Initial experimentation, including preliminary model selection, hyperparameter tuning, and qualitative analysis, was conducted on the Portuguese dataset. This choice was motivated by the team's native proficiency in the language, facilitating a more nuanced understanding of model behavior and output quality before extending the approach to other languages. 

The subsequent subsections detail our data cleaning pipeline, exploratory data analysis insights, the specifics of our modeling techniques, and the evaluation framework.

\subsection{Data Cleaning}
\label{ssec:data_preprocessing}

A recurrent issue observed across multiple languages was the presence of triplicated sentences within original posts, often appended with a \texttt{None} placeholder. This pattern, likely an artifact of automated data collection or formatting, is exemplified in Table~\ref{tab:normalization-example-pt} for a Portuguese instance.

\begin{table}[ht]
\caption{Example of an original Portuguese post exhibiting triplicated content and a trailing \texttt{None}, alongside its normalized version and English translations. The cleaned original post segment (after deduplication by Algorithm~\ref{alg:preprocessing}) is underlined. This pattern was observed across multiple languages.}
\centering
\resizebox{\textwidth}{!}{
\begin{tabular}{p{85mm} p{85mm}}
\toprule
\textbf{Portuguese} & \textbf{English Translation} \\
\midrule
\textbf{Post Original:}  
\ul{Na Holanda, a ministra da Saúde trabalha duas (2) horas diariamente como agente de limpeza antes de ir ao seu escritório. Gostei muito.} Na Holanda, a ministra da Saúde trabalha duas (2) horas diariamente como agente de limpeza antes de ir ao seu escritório. Gostei muito. Na Holanda, a ministra da Saúde trabalha duas (2) horas diariamente como agente de limpeza antes de ir ao seu escritório. Gostei muito. \texttt{None} 
&
\textbf{Original Post:}  
\ul{In the Netherlands, the Minister of Health works two (2) hours daily as a cleaning worker before going to her office. I really liked that.} In the Netherlands, the Minister of Health works two (2) hours daily as a cleaning worker before going to her office. I really liked that. In the Netherlands, the Minister of Health works two (2) hours daily as a cleaning worker before going to her office. I really liked that. \texttt{None} 
\\ 
\textbf{Saída Normalizada:}  
Na Holanda, a ministra da Saúde trabalha duas horas diariamente como agente de limpeza antes de ir ao seu escritório.
&
\textbf{Normalized Claim:}  
In the Netherlands, the Minister of Health works two hours daily as a cleaning worker before going to her office.
\\
\bottomrule
\end{tabular}}%
\label{tab:normalization-example-pt}
\end{table}

To rectify this, we implemented a preprocessing routine (Algorithm~\ref{alg:preprocessing}) designed to first remove any trailing \texttt{None} tokens. Subsequently, it identifies and condenses repeated textual sequences by searching for the smallest repeating pattern that constitutes the entire post. If such a pattern is found, only a single instance of it is retained.

\begin{algorithm}[H] 
\caption{Preprocessing for Repetitive Content and Placeholder Removal}
\label{alg:preprocessing} 
\begin{algorithmic}
\footnotesize
    \Require Raw post $P$
    \Ensure Cleaned post $P_{clean}$

    \State $P_{\text{clean}} \gets P$
    
    \If{$P$ ends with ``None''} \Comment{Remove trailing placeholder}
        \State $P_{\text{clean}} \gets P_{\text{clean}}[:-4].\text{strip}()$ 
    \EndIf
    
    \State $P_{\text{words}} \gets \text{tokenize}(P_{\text{clean}})$ \Comment{Tokenize into words}
    
    \For{$s = 1$ \textbf{to} $\lfloor |P_{\text{words}}|/2 \rfloor$} \Comment{Check for repetitive patterns}
        \State $W \gets P_{\text{words}}[0:s]$ \Comment{Extract candidate pattern}
        \State $\text{num\_repeats} \gets \lfloor |P_{\text{words}}| / s \rfloor$
        \State $\text{repeated\_sequence} \gets W \text{ repeated } \text{num\_repeats} \text{ times}$
        
        \If{$P_{\text{words}}[0 : s \cdot \text{num\_repeats}] = \text{repeated\_sequence}$}
            \State $P_{\text{clean}} \gets \text{join}(W, \text{ spaces})$ \Comment{Return single instance}
            \State \textbf{return} $P_{\text{clean}}$
        \EndIf
    \EndFor
    
    \State \textbf{return} $P_{\text{clean}}$ \Comment{No repetition found}
\end{algorithmic}
\end{algorithm}

Additionally, we implemented cross-split deduplication to handle identical posts appearing in multiple dataset partitions. To preserve evaluation integrity, duplicates were systematically removed by prioritizing retention in test sets, then development sets, and finally training sets.

\subsection{Exploratory Data Analysis (EDA)}
\label{subsec:eda}

\begin{figure}[ht]
    \centering
    \begin{subfigure}[b]{0.495\textwidth}
        \centering
        \includegraphics[width=\linewidth]{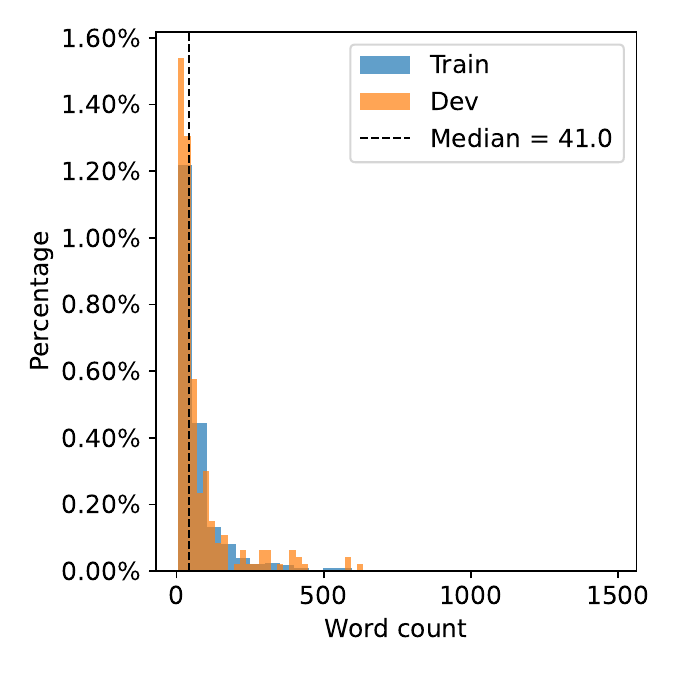}
        \caption{Original posts word count (Portuguese).}
        \label{fig:post_dist}
    \end{subfigure}
    \hfill 
    \begin{subfigure}[b]{0.495\textwidth}
        \centering
        \includegraphics[width=\linewidth]{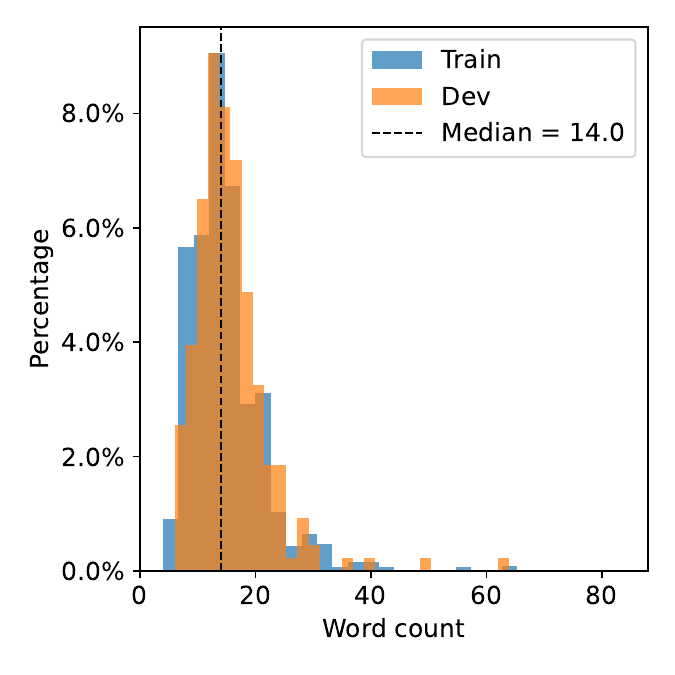}
        \caption{Normalized claims word count (Portuguese).}
        \label{fig:claim_dist}
    \end{subfigure}
    \caption{Histograms of word counts for the Portuguese training and development sets after preprocessing.} 
    \label{fig:combined_dist_charts}
\end{figure}

We analyzed word count distributions for original posts and their corresponding normalized claims within the Portuguese subset following preprocessing (Algorithm \ref{alg:preprocessing}). As depicted for the training and development sets in Figures~\ref{fig:post_dist} and~\ref{fig:claim_dist}, original posts (mean $\approx$ 75 words, STD $\approx$ 107 words; highly skewed) are substantially longer and more variable than normalized claims (mean $\approx$ 15 words, STD $\approx$ 6.8 words). This demonstrates that normalization effectively reduces verbosity and structural noise, leading to more compact and verifiable statements.

The dataset's heterogeneity across languages, as indicated in Table~\ref{tab:splits}, presents further challenges. While some languages offer substantial training samples, others, particularly those in the zero-shot group, provide only minimal test data. This disparity necessitates careful consideration of model generalization and poses a risk of overfitting in resource-rich scenarios and underperformance in low-resource ones.

\section{Experiments}
Our experimental design directly follows the dual-strategy approach detailed in Section~\ref{sec:methodology}, addressing both monolingual (with training and development data) and zero-shot (test data only for specific languages) settings. All input posts were preprocessed according to Algorithm~\ref{alg:preprocessing}.

\subsection{Fine-tuning of Encoder-Decoder SLMs}
This approach involved adapting existing pre-trained encoder-decoder Transformer models for the specific task of claim normalization, utilizing the available training data for the thirteen supervised languages. We primarily sourced these models from the Hugging Face Hub\footnote{\url{https://huggingface.co/models}}. Our strategy prioritized monolingual models, selecting those pre-trained specifically for each target language. The selected monolingual models included:

\begin{itemize}
    \item \textbf{Portuguese:} PTT5 (small, base, large) \citep{carmo2020ptt5pretrainingvalidatingt5}, PTT5-mMARCO (base) \citep{10.5555/3722577.3722647}, PTT5-v2 (small, base, large) \citep{10.1007/978-3-031-79032-4_23},  Mono-PTT5 (small, base, large) \citep{10.1007/978-3-031-79032-4_23}, Portuguese Bart (base) \cite{adalberto_ferreira_barbosa_junior_2024}.
    \item \textbf{Arabic:} AraT5 (base) \citep{nagoudi-etal-2022-arat5}.
    \item \textbf{French:} T5 French (base) \citep{guillaumeT5french}.
    \item \textbf{German}: T5 German (small) \citep{shahmT5German}.
    \item \textbf{Indic languages - Hindi, Marathi, Punjabi, Tamil:}  Varta T5 (base) \citep{aralikatte-etal-2023-varta}.
    \item \textbf{Indonesian:} Indonesian T5 Summarization Base \cite{wirawan2021indonesiant5}. 
    \item \textbf{Polish:} PLT5 (base) \citep{chrabrowa-etal-2022-evaluation}.
    \item \textbf{Spanish:} T5S (base) \citep{araujo-etal-2024-sequence}.
    \item \textbf{Thai:} ThaiT5 Instruct (base) \citep{PeenipatThaiT5Instruct}.
\end{itemize}

In addition to these language-specific models, we also experimented with fine-tuning the following multilingual encoder-decoder architectures on the combined training data of all supervised languages: Flan-T5 (small, base, large) \citep{10.5555/3722577.3722647}, mBART (large) \citep{liu-etal-2020-multilingual-denoising}, and UMT5 (base) \cite{chung2023umt5}.

Fine-tuning was conducted on three distinct hardware platforms: Kaggle kernels equipped with two NVIDIA T4 GPUs, Google Colab Pro sessions with a single NVIDIA T4 GPU, and an on-premise server hosting a single NVIDIA A100 GPU. The A100 GPU, with its substantial memory capacity, was crucial for fine-tuning larger models like PTT5 (Large), which would exceed the memory limits of the T4 GPUs.

\begin{table}[ht]
  \caption{The hyperparameter search space used for fine-tuning. The final configuration for each language was selected based on the best METEOR or BERTScore on the development set.}
  \label{tab:hparams}
  \begin{tabular}{ccl}
    \toprule
    Hyper-parameter & Value \\
    \midrule
    Epochs & \{3, 5, 10, 20\}\\
    Learning rate & \{$3 \times 10^{-3}$, $1 \times 10^{-3}$, $5 \times 10^{-4}$\}\\
    Warm-up steps & 90\\
    Effective Batch size & 32 (via gradient accumulation)\\
    Generation Max Length & \{128, dynamic\}\\
    Optimizer & \{Adafactor, AdamW\}\\ 
    Number of beams & 15 \\
    \bottomrule
  \end{tabular}
\end{table}

We used hyperparameter search space, summarized in Table~\ref{tab:hparams}, in which a gradient accumulation was employed to achieve an effective batch size of 32, even on GPUs with 16 GB of VRAM. The maximum generation length for each batch was dynamically set to the length of the longest target sequence in that batch plus two tokens, minimizing unnecessary padding.

\subsection{Inference with LLMs}
For the monolingual setting, as an alternative to fine-tuning SLMs, we investigated the capabilities of LLMs using few-shot in-context learning in two distinct scenarios: (1) few-shot in-context learning as an alternative to fine-tuning smaller language models (SLMs) in monolingual settings, and (2) zero-shot inference for cross-lingual transfer to languages without available training data. The following LLMs were experimented via their respective APIs: 

\begin{itemize}
    \item \textbf{Google Gemini \cite{gemini2025}:} Gemini 2.0 Flash Lite, Gemini 2.0 Flash Thinking.
    \item \textbf{OpenAI GPT \cite{openai2024gpt4ocard}:} GPT-4o, GPT-4o mini, GPT-4.1 mini.
    \item \textbf{OpenAI Reasoning \cite{openai2024openaio1card, openai2025openaio3card}:} \texttt{o1}, \texttt{o3 mini}.
    \item \textbf{Mistral Pixtral\footnote{\url{https://mistral.ai/news/pixtral-large}}:} Pixtral Large (124B).
    \item \textbf{Alibaba Qwen \cite{hui2024qwen2}:} Qwen 2.5 Instruct (3B).
\end{itemize}

All models were used without post-processing, reclassification, or ensemble methods. Fewer than 0.5\% of API requests returned empty strings, which were retained as-is in our submissions to preserve the authenticity of model outputs.

\subsubsection{Zero-shot Prompting (Zero-shot Setting)}

For zero-shot inference, we developed language-specific prompts that define the normalization task without providing examples. The English prompt (Figure \ref{fig:zero_shot}) served as a template, which was then translated into other languages to ensure consistent task framing. The complete set of prompts is available in Appendix \ref{append:prompts}.

\begin{figure}[ht]
  \centering
  \fbox{\parbox{\textwidth}{You have received an informal and disorganized social media post. Summarize this post into a clear and concise statement, without adding any new information.\\
  \textbf{Post:} \texttt{\{original\_post\}}\\
  \textbf{Normalized statement:}}}
  \caption{The English prompt for zero-shot inference. It instructs the model to normalize the input, inserted at the \texttt{\{original\_post\}} placeholder, by summarizing it concisely while preserving its meaning.}
    \label{fig:zero_shot}
\end{figure}

\subsubsection{Few-shot Prompting (Monolingual Setting)}

In the few-shot setting (applied to monolingual languages as an alternative to SLM fine-tuning), we investigated the impact of varying the number of in-context demonstrations by evaluating prompts with \textbf{3}, \textbf{5}, and \textbf{10 examples} (shots). These examples were selected from the training data of the respective language using several distinct strategies to assess their influence on model performance:

\begin{itemize}
    \item \textbf{Random Selection:} Examples were drawn uniformly at random without replacement.
    \item \textbf{Mixed Difficulty:} Examples combined 'easy' and 'hard' posts (as defined below) to foster robustness.
    \item \textbf{Hard Only:} Examples exclusively featured 'hard' posts (as defined below) to test model performance on challenging cases.
    \item \textbf{HDBSCAN Top-k Prototypes:} Examples comprised prototypes from the $k$ largest HDBSCAN clusters (e.g., $k=3$ or $k=5$), aiming for diverse semantic coverage.
\end{itemize}

For the difficulty-stratified strategies, example difficulty was determined by calculating the METEOR score of the original post against its normalized version for every example in the training dataset. Training examples with the lowest METEOR scores (relative to other examples in the same dataset) were heuristically classified as 'hard', hypothesized to be more challenging for the model or to represent lower-quality reference outputs. Conversely, 'easy' examples were those with the highest METEOR scores.

For HDBSCAN-based strategies, semantically diverse prototypes were selected by first converting posts into sentence embeddings using the `paraphrase-multilingual-MiniLM-L12-v2` model, a distilled Transformer architecture optimized for multilingual sentence-level representations \cite{wang-etal-2021-minilmv2, reimers-2019-sentence-bert}. HDBSCAN (with \texttt{min\_cluster\_size=5}) then clustered these embeddings. The post closest to each cluster's centroid was designated as a prototype. The 'Top-k Prototypes' strategy used prototypes from the $k$ largest clusters, supplemented with random examples if the number of clusters was less than $k$. This approach targets broad semantic coverage with minimal curation.

\section{Results}
\label{sec:results}

This section presents the performance of team AKCIT-FN in the CLEF-2025 CheckThat! Task 2 on Claim Normalization. Our language-adaptive framework achieved podium finishes (top three) in 15 out of the 20 languages evaluated. Specifically, our submissions secured:

\begin{itemize}
  \item \textbf{Second place} in 8 languages: Tamil, Thai, Punjabi, Telugu, Greek, Romanian, Dutch, and Korean.
  \item \textbf{Third place} in 7 languages: Portuguese, Spanish, French, Indonesian, Bengali, Polish, and German.
\end{itemize}

Table~\ref{tab:leaderboard-results} provides a detailed breakdown of our best submission for each language, including the strategy employed, model specifications, average METEOR score, and the official ranking assigned by the organizers.

\begin{table}[ht]
  \centering
  \caption{Comprehensive performance summary showing our best submission per language with strategy, model specifications, METEOR scores, and official rankings. Podium finishes (top 3) are highlighted in bold.} 
  \label{tab:leaderboard-results}
\resizebox{0.75\textwidth}{!}{
\begin{tabular}{@{}clcccc@{}}
\toprule
\multirow{2}{*}{\textbf{Setting}} & \multirow{2}{*}{\textbf{Language}} & \multicolumn{4}{c}{\textbf{Best Submission Details}} \\ \cmidrule(l){3-6} 
 &  & \textbf{Strategy} & \textbf{Model (Parameters)} & \textbf{Avg. METEOR} & \textbf{Rank} \\ \midrule 
\multirow{13}{*}{\makecell[c]{\textbf{Monolingual}\\(training data\\available)}} & \textbf{Portuguese} & \textbf{\makecell{SLM\\Fine-tuning}} & \textbf{\makecell[c]{Mono PTT5 base\\(220M)}} & \textbf{0.5290} & \textbf{3rd} \\
 & \textbf{Spanish} & \textbf{\makecell{SLM\\Fine-tuning}} & \textbf{\makecell[c]{T5S base\\(220M)}} & \textbf{0.5213} & \textbf{3rd} \\
 & \textbf{Tamil} & \textbf{\makecell{SLM\\Fine-tuning}} & \textbf{\makecell[c]{Varta T5 base\\(395M)}} & \textbf{0.5197} & \textbf{2nd} \\
 & English & \makecell{SLM\\Fine-tuning} & \makecell[c]{Flan-T5 base\\(250M)} & 0.4058 & 4th \\
 & \textbf{Indonesian} & \textbf{\makecell{SLM\\Fine-tuning}} & \textbf{\makecell[c]{Indonesian T5 \\ Summarization Base\\(250M)}} & \textbf{0.3866} & \textbf{3rd} \\
 & \textbf{French} & \textbf{\makecell{SLM\\Fine-tuning}} & \textbf{\makecell[c]{T5 French base\\(250M)}} & \textbf{0.3811} & \textbf{3rd} \\
 & Arabic & \makecell{SLM\\Fine-tuning} & \makecell[c]{AraT5 Base\\(220M)} & 0.3277 & 6th \\
 & \textbf{Thai} & \textbf{\makecell{SLM\\Fine-tuning}} & \textbf{\makecell[c]{Thai T5 Base\\(245M)}} & \textbf{0.3179} & \textbf{2nd} \\
 & \textbf{Punjabi} & \textbf{\makecell{SLM\\Fine-tuning}} & \textbf{\makecell[c]{Varta T5 base\\(395M)}} & \textbf{0.3038} & \textbf{2nd} \\
 & \textbf{Polish} & \textbf{\makecell{SLM\\Fine-tuning}} & \textbf{\makecell[c]{plT5 Base\\(275M)}} & \textbf{0.2798} & \textbf{3rd} \\
 & Hindi & \makecell{SLM\\Fine-tuning} & \makecell[c]{Varta T5 base\\(395M)} & 0.2706 & 5th \\
 & \textbf{German} & \textbf{\makecell{SLM\\Fine-tuning}} & \textbf{\makecell[c]{T5 German small\\(60M)}} & \textbf{0.2652} & \textbf{3rd} \\ 
 & Marathi & \makecell{SLM\\Fine-tuning} & \makecell[c]{Varta T5 base\\(395M)} & 0.2181 & 5th \\ \midrule
\multirow{7}{*}{{\makecell[c]{\textbf{Zero-shot}\\(no training\\data)}}} & \textbf{Telugu} & \textbf{\makecell{LLM\\Inference}} & \textbf{\makecell[c]{Qwen 2.5 Instruct\\(3B)}} & \textbf{0.5176} & \textbf{2nd} \\
 & \textbf{Bengali} & \textbf{\makecell{LLM\\Inference}} & \textbf{\makecell[c]{Qwen 2.5 Instruct\\(3B)}} & \textbf{0.2916} & \textbf{3rd} \\
 & \textbf{Greek} & \textbf{\makecell{LLM\\Inference}} & \textbf{GPT-4o Mini} & \textbf{0.2567} & \textbf{2nd} \\
 & \textbf{Romanian} & \textbf{\makecell{LLM\\Inference}} & \textbf{GPT-4o Mini} & \textbf{0.2516} & \textbf{2nd} \\
 & \textbf{Dutch} & \textbf{\makecell{LLM\\Inference}} & \textbf{GPT-4o Mini} & \textbf{0.1922} & \textbf{2nd} \\
 & Czech & \makecell{LLM\\inference} & GPT-4o Mini & 0.1734 & 4th \\
 & \textbf{Korean} & \textbf{\makecell{LLM\\Inference}} & \textbf{GPT-4o Mini} & \textbf{0.1209} & \textbf{2nd} \\
 \bottomrule
\end{tabular}}%
\end{table}

Analysis of our submissions reveals distinct trends based on data availability. 
For the \textbf{monolingual setting} (Table~\ref{tab:leaderboard-results}(a)), where in-domain training data was available, fine-tuned Small Language Models (SLMs) consistently outperformed few-shot Large Language Model (LLM) prompting strategies. This is evidenced by all our best-evaluated submissions in this setting utilizing SLM fine-tuning. 

Notably, the Varta T5 base model (395M parameters) proved highly effective for Indic languages, securing 2nd place for Tamil (0.5197 METEOR) and Punjabi (0.3038 METEOR), and was also the model of choice for Hindi and Marathi (both achieving 5th place). Furthermore, SLM fine-tuning led to 3rd place finishes for Portuguese, Spanish, Indonesian, French, Polish, and German, employing various language-specific T5-based models with parameter counts ranging from 60M (T5 German small) to approximately 275M (PLT5 Base).

Conversely, in the \textbf{zero-shot setting} (Table~\ref{tab:leaderboard-results}(b)), characterized by the absence of language-specific training data, inference with pre-trained Large Language Models (LLMs) demonstrated strong generalization capabilities. Our top performances in this category were achieved using models such as Qwen 2.5 Instruct (3B parameters), which secured 2nd place for Telugu (0.5176 METEOR) and 3rd for Bengali (0.2916 METEOR), and GPT-4o Mini, which achieved 2nd place for Greek, Romanian, Dutch, and Korean. These results underscore the utility of LLMs for rapid adaptation to new languages where specialized training data is scarce.

\section{Conclusion}

This paper detailed AKCIT-FN's participation in the CLEF-2025 CheckThat! Task 2 on multilingual claim normalization. Our approach involved a language-adaptive strategy: fine-tuning language-specific or multilingual Small Language Models (SLMs) for languages with training data, and employing zero-shot prompting with Large Language Models (LLMs) for languages without such data.

Our submissions demonstrated strong performance, achieving podium finishes (top three) in 15 out of the 20 languages. Notably, fine-tuned SLMs excelled in supervised settings, while LLMs proved effective for zero-shot generalization, securing five second-place and one third-place finish among the seven zero-shot languages. For Portuguese, our primary development language, our best system (Mono PTT5 base) ranked third with a METEOR score of 0.5290.

Overall, these results underscore the complementary strengths of SLM fine-tuning when in-domain data is available and the powerful generalization capabilities of LLMs for rapid deployment in zero-shot scenarios for complex NLP tasks like claim normalization. A limitation of our work, however, is the lack of a qualitative error analysis or a discussion of failure cases, which would be a valuable direction for future investigation. Our code and configurations are publicly available to facilitate further research.

\begin{acknowledgments}
This work has been fully funded by the project Computational Techniques for Multimodal Data Security and Privacy supported by Advanced Knowledge Center in Immersive Technologies (AKCIT), with financial resources from the PPI IoT/Manufatura 4.0 / PPI HardwareBR of the MCTI grant number 057/2023, signed with EMBRAPII.
\end{acknowledgments}

\begin{aideclaration}
During the preparation of this work, the author(s) used Gemini 2.5 Pro \cite{gemini2025} (\texttt{gemini-2.5-pro-exp-03-25}), Claude Sonet 4 \cite{anthropic2025sonnet} (\texttt{claude-sonnet-4-20250514}), GPT-4o \cite{openai2024gpt4ocard} (\texttt{gpt-4o-2024-05-13}) in order to: Paraphrase and reword, Improve writing style, Abstract drafting, and Peer review simulation. After using this tool/service, the author(s) reviewed and edited the content as needed and take(s) full responsibility for the publication’s content.
\end{aideclaration}

\bibliography{sample-ceur}

\appendix

\section{Zero-shot prompts}
\label{append:prompts}

This appendix lists the prompts used for the zero-shot claim normalization task. The English prompt in Figure \ref{fig:zero_shot} served as the template and was translated into the seven target languages. The \texttt{\{post\_text\}} placeholder is replaced with the social media post text during inference.

\begin{itemize}
 \setlength\itemsep{5mm}
 \itemindent=-13pt
    \item \textbf{Czech} \\[1mm]
    \fbox{\parbox{420pt}{Dostanete neformální a neuspořádaný příspěvek ze sociálních sítí.
Shrňte jej do jasného a stručného tvrzení, bez přidávání dalších informací. \\
\textbf{Příspěvek:} \texttt{\{post\_text\}}}}
    \item \textbf{Greek} \\[1mm]
    \fbox{\parbox{420pt}{Σου δίνεται μια ανοργάνωτη και ανεπίσημη ανάρτηση στα κοινωνικά δίκτυα.
Περίληψέ την σε μια σαφή και συνοπτική δήλωση, χωρίς να προσθέσεις επιπλέον πληροφορίες.\\
\textbf{Ανάρτηση:} \texttt{\{post\_text\}}}}
    \item \textbf{Dutch} \\[1mm]
    \fbox{\parbox{420pt}{
    Je ontvangt een informeel en ongeorganiseerd bericht op een sociaal netwerk. Vat het samen in een duidelijke en beknopte verklaring zonder extra informatie toe te voegen.\\
    \textbf{berichten:}\texttt{\{post\_text\}}}}
    \item \textbf{Korean} \\[1mm]
    \fbox{\parbox{420pt}{
    비정형적이고 비공식적인 소셜 미디어 게시물이 주어집니다. 이를 명확하고 간결한 주장으로 요약하십시오. 추가 정보는 포함하지 마십시오.\\
    \textbf{게시물:}\texttt{\{post\_text\}}}} \\
    \item \textbf{Romanian} \\[1mm]
    \fbox{\parbox{420pt}{Prompt: Primești o postare informală și dezorganizată de pe o rețea socială.
Rezum-o într-o afirmație clară și concisă, fără a adăuga informații. \\
\textbf{Postare:}  \texttt{\{post\_text\}}}}
    \item \textbf{Tegulu} \\[1mm]
    \fbox{\parbox{420pt}{\telugufont
 మీకు ఒక అసంఘటితమైన, అనౌపచారికమైన సోషల్ మీడియాలో పోస్ట్ ఇవ్వబడుతుంది.
దీనిని స్పష్టమైన మరియు సంక్షిప్తమైన ప్రకటనగా మార్చండి, అదనపు సమాచారం ఇవ్వకుండా. \\
\textbf{పోస్ట్}:\texttt{\{post\_text\}}}} \\
    \item \textbf{Bengali} \\[1mm]
    \fbox{\parbox{420pt}{\bengalifont
 আপনি একটি অগোছালো এবং অনানুষ্ঠানিক সোশ্যাল মিডিয়া পোস্ট পাচ্ছেন।
এটি একটি স্পষ্ট এবং সংক্ষিপ্ত দাবিতে রূপান্তর করুন, কোনো অতিরিক্ত তথ্য ছাড়াই।
\textbf{পোস্ট:} \texttt{\{post\_text\}}}}
\end{itemize}

\end{document}